\def\year{2021}\relax
\documentclass[letterpaper]{article} 
\usepackage{aaai21}  
\usepackage{times}  
\usepackage{helvet} 
\usepackage{courier}  
\usepackage[hyphens]{url}  
\usepackage{graphicx} 
\usepackage{natbib}  
\usepackage{caption} 
\usepackage[subpreambles]{standalone}
\usepackage{tikz}
\usepackage{pgf}
\usepackage{pgfplots}
\usepgfplotslibrary{groupplots}
\usepgfplotslibrary{dateplot}
\usetikzlibrary{patterns}
\usetikzlibrary{shapes.arrows}
\pgfplotsset{compat=newest}
\usepackage{glossaries}
\newacronym{lop}{LOP}{language-oriented programming}
\newacronym{edb}{EDB}{extensional database}
\newacronym{idb}{IDB}{intensional database}
\newacronym{pp}{PP}{probabilistic programming}
\newacronym{ux}{UX}{user experience}
\newacronym{ppl}{PPL}{probabilistic programming language}
\newacronym[longplural={regions of interest}]{roi}{ROI}{region of interest}
\newacronym{cbma}{CBMA}{coordinate-based meta-analysis}
\newacronym{ibma}{IBMA}{image-based meta-analysis}
\newacronym{mri}{MRI}{magnetic resonance imaging}
\newacronym{fmri}{fMRI}{functional magnetic resonance imaging}
\newacronym{dsl}{DSL}{domain-specific language}
\newacronym{snr}{SNR}{signal-to-noise ratio}
\newacronym{bold}{BOLD}{blood-oxygen-level dependent}
\newacronym{sota}{SOTA}{state of the art}
\newacronym{plp}{PLP}{probabilistic logic programming}
\newacronym{nlp}{NLP}{natural language processing}
\newacronym{pdb}{PDB}{probabilistic database}
\newacronym{cpd}{CPD}{conditional probability distribution}
\newacronym{kc}{KC}{knowledge compilation}
\newacronym[longplural={conjunctive queries}]{cq}{CQ}{conjunctive query}
\newacronym[longplural={unions of conjunctive queries}]{ucq}{UCQ}{%
    union of conjunctive queries}
\newacronym{bid}{BID}{block-independent disjoint}
\newacronym{sdd}{SDD}{sentential decision diagram}
\newacronym{psdd}{PSDD}{probabilistic sentential decision diagram}
\newacronym{wmql}{WMQL}{White Matter Query Language}
\newacronym{mcmc}{MCMC}{Markov chain Monte Carlo}
\newacronym{obdd}{OBDD}{ordered binary decision diagram}
\newacronym{dmn}{DMN}{default mode network}
\newacronym{mgu}{MGU}{most general unifier}
\newacronym{mvdb}{MVDB}{markov view database}

\newglossaryentry{tfidf}{
    name=TFIDF,
    description={Term frequency-inverse document frequency}
}

\newglossaryentry{focus}
{
    name=focus,
    description={Coordinates of an activation reported by a neuroimaging study},
    plural=foci
}

\newglossaryentry{non-repeating}
{
    name=non-repeating query,
    description={A query is non-repeating if each relation symbol occurs at
    most once in the query}
}

\newglossaryentry{possible-world}
{
    name=possible world,
    description={TODO}
}

\newglossaryentry{horn-clause}
{
    name=Horn clause,
    description={TODO},
    plural=Horn clauses
}

\usepackage{amsfonts}
\usepackage{mathtools}
\usepackage{bm}
\usepackage{cleveref}
\usepackage{siunitx}

\title{Complex Coordinate-Based Meta-Analysis with Probabilistic Programming}
\author{%
    Valentin Iovene, Gaston Zanitti, Demian Wassermann\\
}
\affiliations{%
    INRIA Saclay, CEA, Université Paris-Saclay, \\
    bat 145, CEA Saclay, 91191 Gif-sur-Yvette, France
}

\begin{document}

\maketitle

\begin{abstract}
    With the growing number of published \gls{fmri} studies, meta-analysis
    databases and models have become an integral part of brain mapping
    research.
    \Gls{cbma} databases are built by extracting both coordinates of reported
    peak activations and term associations using natural language processing
    techniques from neuroimaging studies.
    Solving term-based queries on these databases makes it possible to obtain
    statistical maps of the brain related to specific cognitive processes.
    However, with tools like Neurosynth, only single-term queries lead to
    statistically reliable results.
    When solving complex queries, too few studies from the database contribute
    to the statistical estimations.
    We design a probabilistic \gls{dsl} standing on Datalog and one of its
    probabilistic extensions, CP-Logic, for expressing and solving complex
    logic-based queries.
    We encode a \gls{cbma} database in a probabilistic program.
    Using the joint distribution of its Bayesian network translation, we show
    that solutions of queries on this program compute the right probability
    distributions of voxel activations.
    We explain how recent lifted query processing algorithms make it possible
    to scale to the size of large neuroimaging data, where knowledge
    compilation techniques fail to solve queries fast enough for practical
    applications.
    Finally, we introduce a method for relating studies to terms
    probabilistically, leading to better solutions for two-term \glspl{cq} on
    smaller databases.
    We demonstrate results for two-term \glspl{cq}, both on simulated
    meta-analysis databases and on the widely used Neurosynth database.
\end{abstract}

\glsresetall

\section{Introduction}

The non-invasivity of \gls{fmri} led it to dominate brain mapping research
since the early 1990s \citep{fmridomination}.
In the past three decades, tens of thousands of published studies acquired and
analysed \gls{fmri} signals, producing new understanding of the human brain and
the cognitive function of its different components.
Quickly, the idea of \emph{meta-analysing} this ever-growing amount of
neuroimaging studies flourished.
By gathering and synthesising the findings of a large corpora of neuroimaging
studies, can we derive new knowledge about the brain's mechanics?
Can we study consensus within the cognitive neuroscience community?
The lack of power of neuroimaging studies undermines the reproducibility of
their findings \citep{poldrack_scanning_2017, botvinik-nezer_variability_2020}.
Can we \emph{build} consensus by aggregating results from several underpowered
studies into more robust findings supported by past literature?
Neuroimaging studies traditionally report peak activation coordinates in a
standard stereotactic coordinate system.
This makes it possible to compare them from one study to another.
These coordinates result from statistical hypothesis testing and represent a
condensed synthesis of which regions of the brain are reported as activated by
a given study.
Directly meta-analysing whole-brain unthresholded statistical maps (i.e.\
\emph{image}-based meta-analysis) is known to yield statistically more powerful
results \citep{salimi-khorshidi_meta-analysis_2009}; and the field is moving in
that direction \citep{gorgolewski_neurovaultorg_2015}.
However, most meta-analyses have resorted to \gls{cbma}: the meta-analysis of
these peak activation coordinates.
In the past ten years, an ecosystem of \gls{cbma} databases and tools was
brought to life \citep[e.g.][]{brainmap, neurosynth}, becoming an integral part
of brain mapping research.
Automatic \gls{cbma} databases, like Neurosynth, extract both natural language
processing features and peak activation coordinates from neuroimaging studies.
These tools are used to derive activation patterns
\citep[e.g.][]{wager_fmri-based_2013, cole_global_2012} or reveal meaningful
cognitive processes through reverse inference
\citep[e.g.][]{smallwood_science_2015, seghier_angular_2013,
chang_decoding_2013, andrews-hanna_default_2014}.
With them, researchers can define more robust regions of interest supported by
past literature.

Nonetheless, currently available tools are limited in the complexity of
queries that they can express and solve on \gls{cbma} data.
Neurosynth \citep{neurosynth} presents brain map for single-term queries.
Although technically feasible, term-based \glspl{cq} lead to underpowered
meta-analyses due to the small number of studies matching the queries.
Methods for exploiting \gls{cbma} data have recently been proposed, but they
concern themselves with either developing new procedures for thresholding
statistical brain maps, or integrating spatial priors into probabilistic models
by correlating nearby voxel activations \citep{cbma}.
We look at neuroimaging meta-analysis from a different angle by improving upon
existing \gls{cbma} literature through the development of a \gls{dsl} that
leverages past research on probabilistic logic programming languages and
databases to \emph{formulate and solve more expressive \gls{cbma} queries}.
We believe that, with this approach, more could be wrung out of this type of
data.

Recently, NeuroQuery \citep{neuroquery} produced meta-analyses using
unstructured text-based queries.
By encoding the relationship between terms in a vocabulary using a regularised
linear model, NeuroQuery can produce brain maps for underrepresented terms (few
studies exactly match the term).
However, NeuroQuery's queries are distinct from and harder to interpret than
database queries, which have clear semantics.
Moreover, producing a brain map from studies related to some term $t_1$
\emph{and not related} to some other term $t_2$ is not possible because
NeuroQuery cannot express logic-based queries.
Finally, NeuroQuery is not a probabilistic model that can be plugged into a
richer hierarchical model combining meta-analyses with heterogeneous
modalities, such as neuroanatomical and ontological knowledge.

Since the 1970s, the computer science community has been working on extending
logic programming languages \citep{prolog, abiteboul_foundations_1995} with
probabilistic semantics to represent knowledge uncertainty inherent to
real-world data \citep[reviewed by][]{deraedt}.
A wide variety of efficient approaches to answering questions (queries) from
these programs were developed; alongside seminal theoretical understandings.
\Acrlongpl{dsl} are not new to the cognitive neuroscience community.
The White Matter Query Language \citep{wmql} was developed to help experts
formally describe white matter tracts in a near-to-English syntax.
To the best of our knowledge, applying these techniques to the formulation and
resolution of logic-based queries on probabilistic \gls{cbma} databases has yet
to be attempted.
This approach could make it possible to formulate elaborate hypotheses on the
brain's function and structure and test them against past cognitive
neuroscience literature.

Adopting a language-oriented programming approach, we use probabilistic logic
programming languages to formulate and solve logic-based queries on \gls{cbma}
databases.
This work fits into a broader project to design a \gls{dsl}, coined
\emph{NeuroLang}, for expressing and testing cognitive neuroscience hypotheses
that combine meta-analysis, neuroanatomical and ontological knowledge.
\emph{The work presented here focuses on the probabilistic semantics of
NeuroLang and its application to term-based \gls{cbma} queries.}

Contributions of this work are three-fold.
First, we investigate the feasibility and technicalities of applying
probabilistic logic programming to \gls{cbma}-based brain mapping.
We propose a way to encode a \gls{cbma} database as a probabilistic logic
program based on CP-Logic \citep{cplogic}, on which complex \gls{cbma} queries
can be solved.
We translate this program to an equivalent Bayesian network representation in
order to show that correct answers to probabilistic queries can be derived from
its factorised joint probability distribution.
Second, we explain how leveraging lifted query processing techniques
\citep{lifted_braz, dichotomytheorem} allows us to scale to the large size of
neuroimaging data at the voxel level.
Third, we propose a relaxed modeling of \gls{tfidf} features to better encode
the relationship between terms and studies and show that fewer samples are
needed to solve two-term \glspl{cq} than traditional approaches, on simulated
and real \gls{cbma} databases.

\section{Background}

\subsection{Term-based queries on \gls{cbma} databases}
\label{sec:term-based-queries}

An example of term-based query formulated in plain English is: ``for each
region of the brain, what is the probability that studies associated with both
terms \emph{insula} and \emph{speech} report its activation?''.
The result of term-based queries are used in forward inference to obtain a map
of the brain's activated regions reported by studies matching the query.

A \gls{cbma} database of $N$ studies with a fixed vocabulary of $M$ terms can
be represented as two matrices $\bm{X} \in \mathbb{R}^{N, M}$ and $\bm{Y} \in
\{0, 1\}^{N, K}$, where $\bm{X}_{ij}$ is a \gls{tfidf} feature measured for
term $j$ in study $i$ and $\bm{Y}_{ik} = 1$ if voxel $k$ is reported as
activated in study $i$.
In practice, $\bm{Y}$ is a sparse matrix because only a small proportion of
voxels are reported within a single study.

Forward inference brain maps are constructed from a probabilistic model where
binary random variables $A_k$ and $T_j$ respectively model the activation of
each voxel $v_k$ and the association of studies to each term $t_j$.
$\textbf{P}\left[A_k\middle|T_j\right]$ is the probability that voxel $k$
activates in studies conditioned on the studies being associated with term $j$
and $\textbf{P}\left[A_k\middle|T_\text{insula} \wedge T_\text{speech}\right]$
is the probability that voxel $k$ activates in studies conditioned on studies
being associated with both terms `insula' and `speech'.\footnote{We use
$\textbf{P}\left[A_k\middle|T_i, T_j\right]$ to denote $\textbf{P}\left[A_k =
1\middle|T_i = 1, T_j = 1\right]$.}

Neurosynth \citep{neurosynth} associates terms to studies by applying a
threshold $\tau$ to \gls{tfidf} features $\bm{X}$.
Forward inference maps are obtained by estimating, for each voxel $k$,
\begin{equation}
    \textbf{P}\left[A_k\middle|T_j\right]
    =
    \frac{%
        \sum_{i=1}^N
        \bm{Y}_{ik}
        \bm{1}[\bm{X}_{ij} > \tau]
    }{%
        \sum_{i=1}^N
        \bm{1}[\bm{X}_{ij} > \tau]
    }
    \label{eq:neurosynth-single-term}
\end{equation}
Solving a query with a $p$-term conjunction, $\varphi = T_1 \wedge \cdots
\wedge T_p$, is done by estimating, for each voxel $k$,
\begin{equation}
    \textbf{P}\left[A_k\middle|\varphi\right]
    =
    \frac{%
        \sum_{i=1}^N
        \bm{Y}_{ik}
        \bm{1}[\text{min}(\bm{X}_{i1}, \dots, \bm{X}_{ip}) > \tau]
    }{%
        \sum_{i=1}^N
        \bm{1}[\text{min}(\bm{X}_{i1}, \dots, \bm{X}_{ip}) > \tau]
    }
    \label{eq:neurosynth-two-term}
\end{equation}
As terms are added to this conjunction (and thus, complexity to the query), the
term $\bm{1}[\text{min}(\bm{X}_{i1}, \dots, \bm{X}_{ip}) > \tau]$ goes to
zero for an increasing number of studies.
Rapidly, obtaining a meaningful brain map becomes infeasible due to
statistically weak results.
A different model that relaxes the hard thresholding of \gls{tfidf} features is
proposed in \cref{sec:st}.
Note that, solving a \emph{disjunction} of two terms is done by replacing
$\text{min}$ with $\text{max}$, thereby requiring that only one of the
\gls{tfidf} features passes the threshold.
The more terms are added, the larger the number of studies that are included in
the estimation.
In that case, statistical power is thus not an issue.

\subsection{\Acrlong{plp}}

Before diving into how probabilistic logic programming can be used to encode
\gls{cbma} data, we give a brief introduction to those languages through the
example of CP-Logic.
We also define the syntactic restrictions of the subset of this language that
we use in our \gls{dsl}.

\subsubsection{CP-Logic}

We use CP-Logic \citep{cplogic} as an intermediate representation in the
compilation of our \gls{dsl}.
In CP-Logic, programs contain rules (also called \emph{CP-Events}) of the form
\begin{equation}
    (h_1 : p_1 \vee \cdots \vee h_n : p_n) \gets \varphi
\end{equation}
where $h_i$ are head predicates, $p_i$ are probabilities such that $\sum_i p_i
\leq 1$, and the implication rule's body (also called \emph{antecedent})
$\varphi$ is a first-order logic formula.
All variables occurring in the head (also called \emph{consequent}) of the rule
must also occur in $\varphi$.
Such rules are interpreted as `$\varphi$ being true causes one of the atoms
$h_i$ to be true'.
Which $h_i$ becomes true is drawn from the probability distribution defined by
probabilities $p_i$.
CP-Logic programs define a probability distribution over the set of
\emph{possible worlds} \citep{sato_statistical_1995} associated with possible
executions of the probabilistic program.

\subsubsection{Syntactic restrictions and probabilistic databases}

Only a \emph{subset} of CP-Logic's expressive syntax is necessary to encode a
\gls{cbma} database and formulate term-based queries on it.
In NeuroLang, two kinds of rules are allowed.
Deterministic rules $( h : 1 ) \gets \varphi$, where $\varphi$ is a conjunction
of predicates and $h$ is a single head predicate that is true with probability
$1$ whenever $\varphi$ is true.
Probabilistic rules whose body is $\top$ (always true).
If the head of the rule contains a single predicate, it is a probabilistic
fact.
If it contains more than one head predicate, it is a probabilistic choice.
Moreover, recursive rules such as $(\text{A}(x) : 0.3) \gets A(y) \wedge
\text{B}(x)$ are not permitted in the program.

With these syntactic restrictions, probabilistic rules define relations in a
probabilistic database.
If a rule has more than one head predicate, its tuples are mutually
exclusive and partition the space of possible worlds.
Queries with mutually exclusive predicates are rewritten to be compatible with
probabilistic tuple-independent databases.
Deterministic rules of the program define \emph{\glspl{ucq}} on these
relations.
A \gls{ucq} $Q(\bm{x})$ is defined by a disjunction $\text{CQ}_1(\bm{x}),
\dots, \text{CQ}_n(\bm{x})$, where $\text{CQ}_i(\bm{x})$ are \glspl{cq} which
conjunct logic literals.
One major theoretical result in the field of probabilistic databases is the
\emph{dichotomy theorem} \citep{dichotomytheorem}.
It classifies \glspl{ucq} based on their complexity: those that can be solved
in polynomial time and those that are \#P-hard, in the size of the database.
A set of rules analyses the syntax of a given \gls{ucq} $Q(\bm{x})$ to derive
an algebraic expression that solves $\textbf{P}[Q(\bm{x})]$: the probability of
$Q(\bm{x})$ being true over all possible groundings of the database (i.e.\
possible worlds).
This resolution strategy is called \emph{lifted query processing}.
Guarantees on the efficiency of query resolution is of particular interest in
the context of neuroimaging's high-dimensional space.

\section{Probabilistic \gls{cbma} databases}
\label{sec:pp}

We now describe how \gls{cbma} data and queries can be encoded as a CP-Logic
program.
We then show how this program can be translated to a Bayesian network.
We use its factorised joint probability distribution to analytically derive the
same solutions for term-based queries as in \cref{sec:term-based-queries}.
Finally, we describe our approach to solving queries on this program using
lifted query processing strategies.

\subsection{Encoding a \gls{cbma} database as a probabilistic logic program}

The program of \cref{fig:program} encodes a \gls{cbma} database.
The equiprobable choice on the SelectedStudy relation partitions the space of
possible worlds such that each one corresponds to a particular study.
VoxelReported and TermInStudy predicates encode matrices $\bm{Y}$ and $\bm{X}$.
\begin{figure*}[t]
    \centering
    \begin{tikzpicture}[thick,auto,node distance=1cm]
    \node [] (program) {%
        $\begin{aligned}
            \left(
                \text{TermInStudy}(t_j, s_i)
                : 1
            \right)
            &\gets
            \top
            .
            \quad\quad
            \forall i \in N, \forall j \in M, \bm{X}_{ij} > \tau
            \\
            \left(
                \text{VoxelReported}(v_k, s_i)
                : 1
            \right)
            &\gets
            \top
            .
            \quad\quad
            \forall i \in N, \forall k \in K, \bm{Y}_{ik} = 1
            \\
            \left(
                \bigvee_{i = 1}^N
                \text{SelectedStudy}(s_i)
                : \frac{1}{N}
            \right)
            &\gets
            \top
            .
            \\
            \left(
                \text{TermAssociation}(t)
                : 1
            \right)
            &\gets
            \exists s \left(
                \text{TermInStudy}(t, s) \wedge
                \text{SelectedStudy}(s)
            \right)
            .
            \\
            \left(
                \text{Activation}(v)
                : 1
            \right)
            &\gets
            \exists s \left(
                \text{VoxelReported}(v, s),
                \text{SelectedStudy}(s)
            \right)
            .
        \end{aligned}$
    };
    \begin{pgfonlayer}{background}
    \filldraw [line width=2mm,join=round,black!8]
    (program.north  -| program.west)  rectangle (program.south  -| program.east);
    \end{pgfonlayer}
\end{tikzpicture}
    \caption{%
        CP-Logic program encoding a probabilistic \gls{cbma} database.
        $\text{TermInStudy}(t, s)$ models the presence of term $t$ in study
        $s$.
        $\text{VoxelReported}(v, s)$ encodes whether voxel $v$ was reported in
        study $s$.
        The large $\text{SelectedStudy}$ equiprobable choice over studies makes
        each possible world correspond to a specific study.
        $\text{Activation}(v)$ and $\text{TermAssociation}(t)$ respectively
        model the activation of voxel $v$ and the association with term $t$.
        The SUCC query $\textbf{P} \left[ \text{Activation}(v) \right]$ gives
        the marginal probability of activation of voxels over all studies.
        The query $\textbf{P} \left[ \text{Activation}(t) \middle|
        \text{TermAssociation}(\text{insula}) \right]$ results in a forward
        inference map for the term \emph{insula}.
    }
    \label{fig:program}
\end{figure*}
We write the program such that solving the query $\textbf{P}\left[
    \text{Activation}(v) \middle| \varphi\right]$, where $\varphi$ conjuncts
and/or disjuncts $\text{TermAssociation}(t_j)$ atoms, produces the
probabilistic model of term-based \gls{cbma} queries described in
\cref{sec:term-based-queries}.
For instance, when defining
\begin{equation*}
    \varphi =
    \text{TermAssociation}(\text{insula})
    \wedge
    \text{TermAssociation}(\text{speech})
\end{equation*}
$\textbf{P}\left[\text{Activation}(v)\middle|\varphi\right]$ is
equivalent to the query $\textbf{P}\left[A_k\middle|T_\text{speech} \wedge
T_\text{insula}\right]$ described previously.
We show that in the next section.

\subsection{Equivalence between the program of \cref{fig:program} and the
\gls{cbma} approach of \cref{sec:term-based-queries}}

To justify the design of the program in \cref{fig:program}, we translate it to
an equivalent Bayesian network representation using the algorithm proposed by
\citet{cplogicbn}.
The resulting Bayesian network is depicted in \cref{fig:bn} using
plate-notation.
\begin{figure*}[t]
    \centering\begin{tikzpicture}[thick,auto,node distance=58pt]
    \tikzstyle{n}=[circle,draw=black!75,fill=white,minimum size=10pt,inner sep=0pt]
    \tikzstyle{pil}=[->,thick,shorten <=2pt,shorten >=2pt]

    \node [n] (css) [label=right:$c^\text{SS}$] {};
    \node [n] (ss) [label=right:{$\text{SelectedStudy}(s_i)$}, below of=css,
    yshift=30pt] {};
    \node [n] (and1) [below left of=ss,yshift=20pt,xshift=10pt] {$\bm{\wedge}$};
    \node [n] (and2) [below right of=ss, yshift=20pt,xshift=-10pt] {$\bm{\wedge}$};
    \node [n] (tis)
    [label={below:{$\text{TermInStudy}(t_j, s_i)$}}, left of=and1] {};
    \node [n] (vr) [label={below:{$\text{VoxelReported}(v_k, s_i)$}}, right of=and2] {};
    \node [n] (cji) [label=below:$c_{ji}^\text{TIS}$, left of=tis, xshift=-10pt] {};
    \node [n] (cki) [label=below:$c_{ki}^\text{VR}$, right of=vr, xshift=10pt] {};
    \node [n] (ta) [label=left:$\text{TermAssociation}(t_j)$, below
    of=and1,yshift=23pt] {$\bm{\vee}$};
    \node [n] (a) [label=right:$\text{Activation}(v_k)$, below
    of=and2,yshift=23pt] {$\bm{\vee}$};

    \node (whatever1) [below of=cki,yshift=50pt] {};
    \node (whatever2) [below of=ss,yshift=24pt]
    {\textbf{$i \in N$}};
    \node (whatever3) [above of=cji,yshift=-54pt] {};
    \node (whatever4) [above of=cki,yshift=-54pt] {};

    \node (whatever5) [right of=a,xshift=60pt]
    {\textbf{$k \in K$}};
    \node (whatever6) [left of=ta,xshift=-60pt]
    {\textbf{$j \in M$}};

    \path[pil] (css) edge (ss);
    \path[pil] (ss) edge (and1);
    \path[pil] (ss) edge (and2);
    \path[pil] (cji) edge (tis);
    \path[pil] (tis) edge (and1);
    \path[pil] (cki) edge (vr);
    \path[pil] (vr) edge (and2);
    \path[pil] (and1) edge (ta);
    \path[pil] (and2) edge (a);

    \node [n] (and-explain) [below of=tis,yshift=-25pt,xshift=-10pt,
    label=below:{\footnotesize$\textbf{P}\left[Y = \top \middle| X_1,\dots,X_n\right] = \prod_{i=1}^N \bm{1}[X_i =
    \top]$},label=left:$Y$]
    {$\bm{\wedge}$};

    \node (whatever8) [above of=and-explain,yshift=-30pt] {\textbf{AND} node};

    \node [n] (and-explain-par-1)
    [above left
    of=and-explain,yshift=-25pt,xshift=25pt,label=left:{$X_1$},]
    {};
    \node [n] (and-explain-par-n)
    [above right
    of=and-explain,yshift=-25pt,xshift=-25pt,label=right:{$X_N$}]
    {};

    \node (whatever7) [above of=and-explain,yshift=-42pt,xshift=1pt]
    {$\cdots$};

    \path[pil] (and-explain-par-1) edge (and-explain);
    \path[pil] (and-explain-par-n) edge (and-explain);

    \node [n] (nor-explain) [below of=vr,yshift=-27pt,xshift=8pt,
    label=below:{\footnotesize$\textbf{P}\left[Y = \top \middle| X_1, \dots, X_n\right] = 1 - \prod_{i=1}^n (1 - \bm{1}[X_i = \top])$},label=right:$Y$]
    {$\bm{\vee}$};

    \node (whatever8) [above of=nor-explain,yshift=-30pt] {\textbf{NoisyOR} node};

    \node [n] (nor-explain-par-1)
    [above left of=nor-explain,yshift=-25pt,xshift=25pt,label=left:{$X_1$}]
    {};
    \node [n] (nor-explain-par-n)
    [above right
    of=nor-explain,yshift=-25pt,xshift=-25pt,label=right:{$X_N$}]
    {};

    \node (whatever7) [above of=nor-explain,yshift=-42pt,xshift=1pt] {$\cdots$};

    \path[pil] (nor-explain-par-1) edge (nor-explain);
    \path[pil] (nor-explain-par-n) edge (nor-explain);

    \matrix[
    matrstandard,
    nodes={draw=black!5, nodostandard, anchor=base west},
    below of=whatever2,
    yshift=-100pt,
    column 1/.style={nodes={text width=10em}},
    column 2/.style={nodes={text width=5em}},
    column 3/.style={nodes={text width=24em}},
    row 1/.style={nodes={fill=black!5}},
    ]
    {%
        \textbf{Random variables} & \textbf{Domain} & \textbf{Conditional
            probability distribution} \\
        $c^\text{SS}$ & $\{0, \dots, N\}$ & $\forall i \in \{1, \dots N\},
        \textbf{P}\left[c^\text{SS} = i\right] = \frac{1}{N}$ \\
        $\text{SelectedStudy}(s_i)$ & $\{ \top, \perp \}$ &
        $\textbf{P}\left[\text{SelectedStudy}(s_i) = \top \middle| c^\text{SS} \right] = \bm{1}[c^\text{SS} = i]$ \\
        $c_{ji}^\text{TIS}$ & $\{0, 1\}$ & $\textbf{P}\left[c_{ji}^\text{TIS} =
        1\right] = \omega(\bm{X}_{ji}; \alpha, \tau)$ \\
        $\text{TermInStudy}(s_i)$ & $\{ \top, \perp \}$ &
        $\textbf{P}\left[ \text{TermInStudy}(s_i) = \top \middle|
        c^\text{TIS} \right] = \bm{1}[c^\text{TIS} = 1]$ \\
        $c_{ki}^\text{VR}$ & $\{0, 1\}$ & $\textbf{P}\left[c_{ki}^\text{VR} =
        1\right] = \bm{Y}_{ki}$ \\
        $\text{VoxelReported}(s_i)$ & $\{ \top, \perp \}$ &
        $\textbf{P}\left[\text{VoxelReported}(s_i) = \top \middle|
        c^\text{VR} \right] = \bm{1}[c^\text{VR} = 1]$ \\
    };

    \begin{pgfonlayer}{background}
        \begin{scope}[
            rounded corners=5pt,
            nw/.style={xshift=-10pt,yshift=10pt},
            se/.style={xshift=10pt, yshift=-10pt},
            snw/.style={xshift=-5pt,yshift=3pt},
            sse/.style={xshift=5pt, yshift=-5pt}]
            \fill [black, opacity=0.07]
            ([nw]{ss.north -| cji.west}) rectangle
            ([se]{whatever1.south -| cki.east});
            \fill [black, opacity=0.07]
            ([snw]{cji.north -| cji.west}) rectangle
            ([sse]{ta.south -| ta.east});
            \fill [black, opacity=0.07]
            ([snw]{cki.north -| and2.west}) rectangle
            ([sse]{a.south -| cki.east});
        \end{scope}
    \end{pgfonlayer}

\end{tikzpicture}
    \caption{%
        Plate-notation representation of the Bayesian network translated from
        the program described in \cref{sec:pp}.
        Each ground atom in the program (e.g. $\text{TermInStudy}(t_2,
        s_{21})$) becomes a binary random variable with a deterministic
        \gls{cpd}.
        Specific \textbf{AND} nodes encode the conjunctions in the antecedent
        of the rules of the program.
        Choice random variables $c^\text{SS}$, $c_{ji}^\text{TIS}$,
        $c_{ki}^\text{VR}$ represent probabilistic choices in the program.
    }
    \label{fig:bn}
\end{figure*}
To simplify the notation, we use $A_k$, $T_n$ and $T_m$ to denote random
variables $\text{Activation}(v_k)$, $\text{TermAssociation}(t_n)$, and
$\text{TermAssociation}(t_m)$.
From the joint probability distribution defined by the Bayesian network, it can
be derived that
\begin{align}
    &\textbf{P}[A_k, T_n, T_m] \\
    &=
    \sum_{i=1}^N
    \textbf{P}[c^\text{SS} = i]
    \textbf{P}[c_{ki}^\text{VR} = 1]
    \textbf{P}[c_{ni}^\text{TIS} = 1]
    \textbf{P}[c_{mi}^\text{TIS} = 1] \\
    &=
    \frac{1}{N}
    \sum_{i=1}^N
    \bm{Y}_{ik}
    \bm{1}[\bm{X}_{in} > \tau]
    \bm{1}[\bm{X}_{im} > \tau]
\end{align}
and, similarly, that
\begin{align}
    \textbf{P}[T_n, T_m]
    &=
    \sum_{i=1}^N
    \textbf{P}[c^\text{SS} = i]
    \textbf{P}[c_{ni}^\text{TIS} = 1]
    \textbf{P}[c_{mi}^\text{TIS} = 1] \\
    &=
    \frac{1}{N}
    \sum_{i=1}^N
    \bm{1}[\bm{X}_{in} > \tau]
    \bm{1}[\bm{X}_{im} > \tau]
\end{align}
From these two joint probability distributions, the solution of the conditional
query can be derived using that
$\textbf{P}[A_k|T_n, T_m] = \frac{\textbf{P}[A_k, T_n, T_m]}{\textbf{P}[T_n,
T_m]} $, which gives the formula of \cref{eq:neurosynth-two-term}, for $p = 2$.

The same can be shown for disjunctive queries $\textbf{P}\left[A_k\middle|T_n
\vee T_m\right]$ by summing the results of 3 two-term \glspl{cq} as follows
\begin{equation}
    \begin{split}
        \textbf{P}\left[A_k\middle|T_n \vee T_m\right]
        &=
        \textbf{P}\left[A_k\middle|T_n, T_m\right] + \\
        & \textbf{P}\left[A_k\middle|\neg T_n, T_m\right] +
        \textbf{P}\left[A_k\middle|T_n, \neg T_m\right]
    \end{split}
\end{equation}

This confirms that the probabilistic program of \cref{fig:program} is sound, as
solving queries on the program leads to the statistical estimation described in
the previous section.

\subsection{Solving queries on probabilistic \gls{cbma} databases}

We now explore query resolution techniques that scale to the size of large
probabilistic \gls{cbma} databases.
The estimation of a forward inference brain map for a two-term conjunction
corresponds to the query
\begin{equation*}
    \footnotesize
    \textbf{P}\left[\text{Activation}(v)\middle|\text{TermAssociation}(t_i),
    \text{TermAssociation}(t_j)\right]
\end{equation*}
We solve this task by defining two \glspl{cq}
\begin{align*}
    Q_1(v) & \gets
    \begin{aligned}
        & \text{Activation}(v), \text{TermAssociation}(t_i), \\
        & \text{TermAssociation}(t_j)
    \end{aligned} \\
    Q_2 & \gets
        \text{TermAssociation}(t_i), \text{TermAssociation}(t_j)
\end{align*}
such that $\frac{\textbf{P}[Q_1(v)]}{\textbf{P}[Q_2]}$ solves the initial
query.
The numerator corresponds the joint probability of a voxel activation and the
association to the two terms. The denominator corresponds to the joint
probability of the associations to the two terms.

\subsubsection{%
    \Gls{kc} approaches do not scale to the size of neuroimaging data
}

We implemented the program of \cref{fig:program} in ProbLog2 \citep{problog2}.
We observed that, when solving two-term \glspl{cq}, grounding and compiling the
program to \glspl{sdd} was impractical.
Solving a two-term \gls{cq} takes more than 30 minutes on a recent laptop.
This is due to the large number of voxels, terms and studies modeled in the
program, leading to a large number of ground literals.
To give perspective on the scale of \gls{cbma} and neuroimaging data, a brain
is typically partitioned into a grid of about \num[group-separator={,}]{230000}
$2\text{mm}^3$ voxels.
On average, studies in the Neurosynth database report
$\num[group-separator={,}]{3165}$ voxel activatons.
There are $\num[group-separator={,}]{14371}$ studies and
$\num[group-separator={,}]{3228}$ terms in the Neurosynth database.
We also tried compiling our program manually to \glspl{sdd} \citep{sdd}.
Despite our efforts, which did note include exploring recent tree-building
strategies~\citep{amarilli:hal-01336514}, the resolution of queries was still
too slow to be practical for real-world applications.
Currently available \gls{cbma} tools are capable of solving single-term queries
in seconds.
Resolution of more complex queries should have a similar time complexity.

\subsubsection{Lifted processing of \glspl{ucq} on probabilistic \gls{cbma}
databases}
\label{sec:lifted}

We leverage theoretical results which have identified classes of queries that
lifted inference can solve in polynomial time.
The dichotomy theorem \citep[proven in][]{dichotomytheorem} provides a
procedure for checking that \glspl{ucq} are \emph{liftable}.
This theorem is convenient because it guarantees that any query such that the
lifted processing rules apply is guaranteed to be solvable in PTIME.
If the query is not liftable, we resort to \gls{kc}-based resolution
techniques.
Because the language does not have probabilistic clauses and prevents
recursivity, we can use its deterministic rules to construct \glspl{ucq}
associated with a given probabilistic query $\textbf{P}[\psi(\bm{x})]$, where
$\psi(\bm{x})$ is a conjunction of intensional, extensional or probabilistic
literals.
This lifted approach makes it possible to solve \gls{cq} in a few seconds.
\emph{Extensional query plans} \citep[see 4.1 of][]{queryprocessing} are
obtained and evaluated to solve queries using a custom Python relational
algebra engine.

\section{Relating terms and studies probabilistically}
\label{sec:st}

The hard thresholding $\bm{1}[x > \tau]$ of \gls{tfidf} features $x$ presented
in \cref{sec:term-based-queries} misses studies that could be relevant to the
resolution of queries.
Because we are interested in solving more complex queries, in this section we
explore a relaxation by introducing the soft-thresholding function
\begin{equation}
    \omega(x ; \alpha, \tau)
    \coloneqq
    \sigma \left( \alpha ( x - \tau ) \right)
    \quad
    \in [0, 1]
    \label{eq:omega}
\end{equation}
where $\sigma$ is the logistic function and $\tau$ a threshold.
As $\alpha$ increases, $\omega( x ; \alpha, \tau )$ converges towards the
hard-thresholding function $\bm{1}[x > \tau]$.
With an appropriate $\alpha$, a larger proportion of studies is included in the
calculation of $\textbf{P}\left[A_k\middle|\varphi\right]$, giving better
estimates on small databases.
For example, results of two-term \gls{cq} $\textbf{P}\left[A_k\middle|T_1
\wedge T_2\right]$ and \gls{ucq} $\textbf{P}\left[A_k\middle|T_1 \vee
T_2\right]$ queries are estimated with
\begin{align}
    \textbf{P}\left[A_k\middle|T_1 \wedge T_2\right]
    &=
    \frac{%
        \sum\limits_{i = 1}^N
        \bm{Y}_{ik}
        \omega\left(
            \bm{X}_{i1} ; \alpha, \tau
        \right)
        \omega\left(
            \bm{X}_{i2} ; \alpha, \tau
        \right)
    }{%
        \sum\limits_{i = 1}^N
        \omega\left(
            \bm{X}_{i1} ; \alpha, \tau
        \right)
        \omega\left(
            \bm{X}_{i2} ; \alpha, \tau
        \right)
    }\label{eq:conjunction-st}
    \\
    \textbf{P}\left[A_k\middle|T_1 \vee T_2\right]
    &=
    \frac{%
        \sum\limits_{i = 1}^N
        \bm{Y}_{ik}
        \big(
            1 -
            \prod\limits_{j=1}^2
            \left( 1 -
                \omega\left(
                    \bm{X}_{ij} ; \alpha, \tau
                \right)
            \right)
        \big)
    }{%
        \sum\limits_{i = 1}^N
        \big(
            1 -
            \prod\limits_{j=1}^2
            \left( 1 -
                \omega\left(
                    \bm{X}_{ij} ; \alpha, \tau
                \right)
            \right)
        \big)
    }\label{eq:disjunction-st}
\end{align}
More generally, $\textbf{P}\left[A_k\middle|\varphi\right]$ can be estimated
for first-order logic formulas $\varphi$ that blend conjunctions and
disjunctions of Boolean random variables $T_j, j \in 1, \dots, M$.
For example, if $\varphi = (T_1 \vee T_2) \wedge (T_3 \vee T_4)$, we have
\begin{equation}
    \textbf{P}\left[A_k\middle|\varphi\right]
    =
    \frac{%
        \sum\limits_{i = 1}^N
        \bm{Y}_{ik}
        f(\bm{X}_{i1}, \bm{X}_{i2})
        f(\bm{X}_{i3}, \bm{X}_{i4})
    }{%
        \sum\limits_{i = 1}^N
        f(\bm{X}_{i1}, \bm{X}_{i2})
        f(\bm{X}_{i3}, \bm{X}_{i4})
    }
\end{equation}
where
$f(x_1, x_2)
=
1 -
(1 - \omega(x_1 ; \alpha, \tau))
(1 - \omega(x_2 ; \alpha, \tau))
$.

This modeling is implemented simply by integrating $\omega(\bm{X}_{ij} ;
\alpha, \tau)$ as the probabilities of probabilistic facts
$\text{TermInStudy}(t_j, s_i)$ in the program of \cref{fig:program}.

\section{Experiments and results}

We compare our method with Neurosynth's on simulated \gls{cbma} databases
sampled from a generative model and on the Neurosynth database.
Using both models, we solve 55 different two-term \glspl{cq}
$\textbf{P}\left[A_k\middle|T_i \wedge T_j\right]$.

\subsubsection{%
    Gain of statistical power when solving two-term \glspl{cq}
    $\textbf{P}\left[A_k\middle|T_i \wedge T_j\right]$ on smaller simulated
    \gls{cbma} databases
}
\label{sec:simulations}

\begin{figure}
    \centering
    \begin{tikzpicture}[thick,auto,node distance=0.85cm]
    \tikzstyle{n}=[circle,draw=black!75,fill=white,minimum size=10pt,inner sep=0pt]
    \tikzstyle{pil}=[->,thick,shorten <=2pt,shorten >=2pt]

    \node [n] (ztfidf) [label=right:{$Z_\text{TFIDF}^{(i)} =
        Z_\text{TF}^{(i)} \times Z_\text{IDF}$}] {};
    \node [n] (zidf) [left of=ztfidf, label=left:{$Z_\text{IDF}$}] {};
    \node [n] (ztf) [above of=ztfidf,label=right:{$Z_\text{TF}^{(i)} \sim
        \sigma\left(\mathcal{N}(\bm{\mu}, \bm{\Sigma})\right)$}] {};
    \node [n] (pki) [below of=ztfidf,label=right:{%
        $p_k^{(i)} = \sigma( \beta_k \cdot Z_\text{TFIDF} )$
    }] {};
    \node [n] (bk) [left of=pki,label=below:$\beta_k$] {};
    \node [n] (Ak) [below of=pki,label=right:$A_k^{(i)} \sim \mathcal{B}(p_k)$] {};

    \node (whatever) [right of=pki] {};
    \node (whatever2) [right of=whatever] {};
    \node (whatever3) [right of=whatever2,xshift=2pt] {};
    \node (whatever4) [right of=whatever3,xshift=-30pt] {};

    \node (plate) [below of=whatever4,xshift=13pt,yshift=-2pt]
    {\textbf{\scriptsize$k \in V$}};
    \node (plate2) [below of=plate,yshift=11pt,xshift=8pt]
    {\textbf{\scriptsize$i \in N$}};

    \path[pil] (ztf) edge (ztfidf);
    \path[pil] (zidf) edge (ztfidf);
    \path[pil] (ztfidf) edge (pki);
    \path[pil] (ztf) edge (zidf);
    \path[pil] (bk) edge (pki);
    \path[pil] (pki) edge (Ak);

    \node (logistic) [minimum size=20pt, below left
    of=Ak,xshift=-5pt,yshift=-15pt,label={[label distance=-8pt]below:{\scriptsize~logistic}}] {$\sigma$};

    \node (bernoulli) [minimum size=20pt, right of=logistic,xshift=20pt,label={[label
        distance=-8pt]below:{\scriptsize~bernoulli}}] {$\mathcal{B}$};

    \node (normal) [minimum size=20pt, right of=bernoulli,xshift=20pt,label={[label
        distance=-8pt]below:{\scriptsize~gaussian}}] {$\mathcal{N}$};

    \begin{pgfonlayer}{background}
        \begin{scope}[
            rounded corners=5pt,
            nw/.style={xshift=-5pt,yshift=5pt},
            se/.style={xshift=25pt, yshift=-5pt},
            snw/.style={xshift=-5pt,yshift=3pt},
            sse/.style={xshift=1pt, yshift=-1pt}]
            \fill [black, opacity=0.06]
            ([nw]{pki.north -| bk.west}) rectangle
            ([se]{Ak.south -| whatever4.east});
            \fill [black, opacity=0.06]
            ([snw]{ztf.north -| ztf.west}) rectangle
            ([sse]{plate2.south -| plate2.east});
        \end{scope}
    \end{pgfonlayer}
\end{tikzpicture}
    \caption{%
        Model for generating \gls{cbma} databases of size $N$.
        $Z_\text{TF}^{(i)}$ models term frequencies in study $i$ and
        follows a logistic-normal distribution.
        $Z_\text{IDF}$ computes inverse document frequencies from
        $\{Z_\text{TF}^{(i)}\}_{i \in N}$.
        $p_k$ is the probability of activation of voxel $v_k$.
        Vectors $\beta_k$ are obtained from a rejection sampling
        scheme that controls the proportion of voxels that activate when
        the query is verified.
        $Z_\text{IDF}$, $\bm{\mu}$ and $\bm{\Sigma}$ are estimated from
        4168 scrapped PubMed abstracts.
    }
    \label{fig:generative-model}
\end{figure}
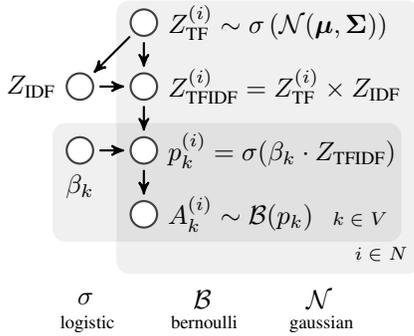

\begin{figure*}[t]
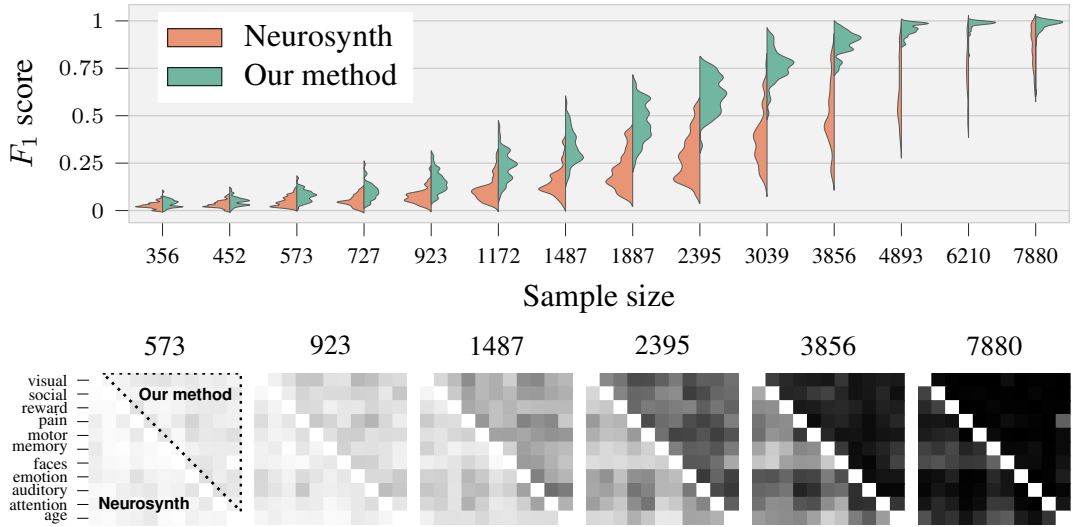

    \centering\includestandalone[width=0.8\textwidth]{fig_simulation_f1_score}
    \centering\includestandalone[width=0.8\textwidth]{fig_simulation_f1_mats}
    \caption{%
        Comparison of Neurosynth's and our method's $F_1$ scores across 55
        \glspl{cq} $\textbf{P}\left[A_k\middle|T_i \wedge T_j\right]$
        on simulated \gls{cbma} databases of varying sample sizes.
        For each sample size, 100 random sub-samples were used.
        \textbf{Above}, $F_1$ score distributions on all queries are compared
        across sample sizes.
        \textbf{Below}, $F_1$ score matrices (white is $0$, black is $1$) are
        compared across sample sizes.
        The upper triangular contains scores of our method and the lower
        triangular contains scores of Neurosynth.
        The threshold $\tau = 0.1$ is used in both models.
        The value $\alpha = 300$ was empirically chosen.
        Varying $\alpha$ near this value does not change the results
        noticeably.
        Sample sizes were taken on a logarithmic scale.
    }\label{fig:simulation}
\end{figure*}

We evaluate our method on simulated small \gls{cbma} databases obtained by
sampling from the generative model of \cref{fig:generative-model}.
This generative model provides the ground truth of which voxels activate in
studies matching a given query of interest.
This binary classification setting makes it possible to compare models by
measuring their ability to identify true voxel activations for multiple sample
sizes.
We experimented with multiple numbers of voxels ($K \in [100, 1000]$).
Preliminary results showed that varying the number of voxels in this range does
not alter the results.
We report results for $K = \num[group-separator={,}]{1000}$ voxels, of which
5\% are activated in studies matching the query.
Predicted voxel activations are obtained by thresholding $p$-values computed
from each model's estimation of $\textbf{P}\left[A_k\middle|\varphi\right]$
using a $G$-test of independence.
We use a $p$-value threshold of $0.01$ and a Bonferroni correction for multiple
comparisons.
Simulation results for two-term \glspl{cq} are presented in
\cref{fig:simulation}, where we compare our model's and Neurosynth's $F_1$
scores across 55 two-term \glspl{cq}.
These queries correspond to all two-term combinations out of 11 terms (depicted
on the $y$-axis of the bottom plot of \cref{fig:simulation}) associated with a
sufficiently large number of studies within the Neurosynth database to produce
meaningful forward inference map.
The $F_1$ score measures the performance of a binary classifier by combining
its precision and recall into a single metric.
We see the advantage of our approach over Neurosynth's for smaller generated
samples where activations related to the query can be identified more reliably
(higher $F_1$ scores).
Multiple values of $\alpha$ in the range $[100, 1000]$ were tried during our
experiments.
However when $\alpha$ is too small or too large, the model tends to include
either too many (and irrelevant) or too few studies in the estimation.
When $\alpha$ tends to $0$, $\omega$ becomes equivalent to Neurosynth's hard
thresholding.
We found that a sweet spot for $\alpha$ was around $300$ and report results for
that value.
Drawing the sigmoid curve for $\alpha = 300$ confirms that this transformation
of TFIDF features is adequate because it maintains Neurosynth's hard
thresholding's property of giving a $0$ or $1$ probability to the lowest and
highest TFIDF features (respectively).

The proposed approach did not show an advantage over Neurosynth for solving
two-term disjunctive queries.
This is expected, as such queries do not reduce the number of studies
incorporated in the estimation of $\textbf{P}\left[A_k\middle|T_i \vee
T_j\right]$, as explained in \cref{sec:term-based-queries}.

\subsubsection{Gain of activation consistency on a real \gls{cbma} database}
\label{sec:real}

We evaluate our method on the Neurosynth \gls{cbma} database.
Because we don't have a ground truth of which voxels activate for a given
query, we resort to comparing models based on the \emph{consistency} of their
predicted activations over many random sub-samples of the Neurosynth database.

From predicted activation maps of $K$ voxels obtained from $M$ sub-samples of a
\gls{cbma} database, the \emph{consistency} for a two-term conjunctive formula
$\varphi$ is computed as
\begin{equation}
    C_\varphi
    \coloneqq
    \frac{1}{K}
    \sum_{k = 1}^K
    \left(
    1 - 2 \times \left|
    \frac{1}{M}
    \sum_{m = 1}^M
    \hat{y}_{mk}^\varphi
     - \frac{1}{2}
    \right|
    \right)
\end{equation}
where $\hat{y}_{mk}^\varphi = 1$ if voxel $k$ is predicted to be activated in
sub-sample $m$ when formula $\varphi$ is true.
The closer to one, the closer the average activation is to $0$ or $1$, which
indicates a higher consistency across sub-samples.
The closer to zero, the closer the average activation is to $0.5$ which
indicates that the predicted activations are highly variable across samples.

\begin{figure}[t]
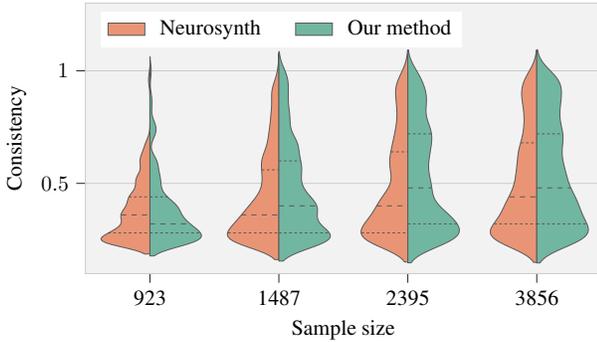

    \centering
    \includestandalone[width=0.45\textwidth]{fig_real_result}
    \caption{%
        Comparison of both models' distributions of voxel activation
        consistency across \num[group-separator={,}]{1000} sub-samples of the
        Neurosynth's database, for 55 two-term \glspl{cq} and for multiple
        sample sizes.
        As the sample size increases, our method finds more consistent
        activations than Neurosynth.
    }\label{fig:real}
\end{figure}

Results are reported in \cref{fig:real}, where the distribution of
consistencies across the same 55 \glspl{cq} as in
\cref{sec:simulations} are shown for multiple sample sizes.
For the largest sample sizes, consistency scores are closer to $1$ with our
method than with Neurosynth's.
For a sample size of 2395 (chosen on a logarithmic scale), the average
consistency of our method was 0.48 while Neurosynth's was 0.4 (+20\%) across
samples and queries. For a sample size of 3856, we notice a 10\% improvement.

We did not experiment with larger sample sizes due to the computational cost of
running the experiment on many Neurosynth subsamples for all \glspl{cq}.
Also, we were mainly interested in whether our approach would be more
consistent for smaller sample sizes.
We observed that the consistency between Neurosynth and our approach was
similar when both models were estimated on the entire database.
This means that the proposed approach is more consistent on smaller sample
sizes but equivalently consistent on larger sample sizes.
The NeuroLang program implementing this model is available at
\url{https://github.com/NeuroLang/NeuroLang/tree/master/examples/plot_neurosynth_relaxed_tfidf.py}.

\section{Discussion}

This work fits into a broader approach to design a \acrlong{dsl} for expressing
logic-based cognitive neuroscience hypotheses that combine neuroimaging data,
neuroanatomical probabilistic maps, ontologies and meta-analysis databases to
produce fine-grained brain maps supported by past literature and heterogeneous
data.

The number of voxels ($K = \num[group-separator={,}]{1000}$) used in the
simulations of \cref{sec:simulations} is orders of magnitude lower than on the
typical whole-brain neuroimaging setting, where $K \simeq \mathcal{O}(10^5)$.
We chose to lower the dimension in the simulation setting for computational
practicality purposes.
However, we believe that maintaining the same proportion of reported
activations as in the Neurosynth database was enough to confirm our approach on
simulations before applying it to real data, as reported in \cref{sec:real}.

The flexible syntax of logic-based language allow to express all kinds of
queries.
$\textbf{P}\left[{\text{TermAssociation}(t)}\middle|{\varphi_\text{insula} \vee
\varphi_\text{fMRI}}\right]$ queries for terms that are most probably
associated with a given pattern of activation, where $\varphi_\text{insula}$ is
a conjunction of logic predicates $\text{Activation}(v_k)$ whose probabilities
come from a neuroanatomical probabilistic map of the insula, and where
$\varphi_\text{fMRI}$ is also a conjunction of predicates
$\text{Activation}(v_k)$ whose voxels $v_k$ are based on neuroimaging data
coming from a custom \gls{fmri} study.

The current version of NeuroLang is limited in what it can model, mainly due
to the syntactic limitations on programs and queries that we had to make in
order to use lifted processing strategies that scale to the size of \gls{cbma}
data.
In cognitive neuroscience, there is interest in using spatial priors to give
the incentive to nearby voxels to co-activate \citep{kong_spatial_2018}.
Spatial priors could be formulated as recursive probabilistic rules, such as
$\text{Activation}(v_1) : f(d) \gets \text{Activation}(v_2),
d = \texttt{euclidean}(v_1, v_2)$, where $d$ is the Euclidean distance measure
between two regions of the brain, and where $f$ maps $d$ to a proper
probability in $[0, 1]$.
The resolution of such queries remains a challenge both in terms of methodology
and tractability.
Future progress in the field of probabilistic programming languages could open
the door to other queries of interest to the cognitive neuroscience community.

\section{Conclusion}

This work is a step towards incorporating complex meta-analyses in brain
mapping models.
We encode a \gls{cbma} database in a probabilistic logic program on which
general logic-based queries can be solved.
Leveraging efficient query resolution strategies on probabilistic databases, we
are able to scale to the size of neuroimaging data.
We experimented with a new method for solving two-term \glspl{cq} using
\gls{tfidf} features more efficiently than the hard thresholding scheme used by
Neurosynth.
This is promising but further investigation should be conducted to know whether
this method extends to queries that conjunct more terms or queries that blend
conjunctions and disjunctions.
The proposed method requires the same computational power as Neurosynth.

\section*{Free software}

The source code of NeuroLang is freely available on GitHub%
\footnote{\url{https://github.com/NeuroLang/NeuroLang}}.

\section*{Acknowledgments}

This work was funded by the ERC-2017-STG NeuroLang grant.
We thank the AAAI-2021 reviewers and the senior program committee member for
the quality and rigor of their comments on this work.
We also thank the organisers of the conference for orchestrating the review of
such a large number of submissions.
We are grateful to some of our colleagues who shared insights on our work:
Antonia Machlouzarides-Shalit, Hicham Janati, and Louis Rouillard-Odera.

\bibliography{bibliography}

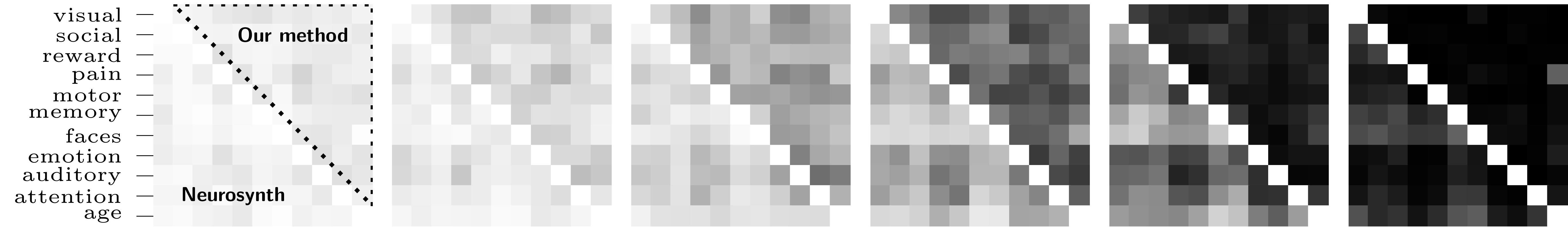
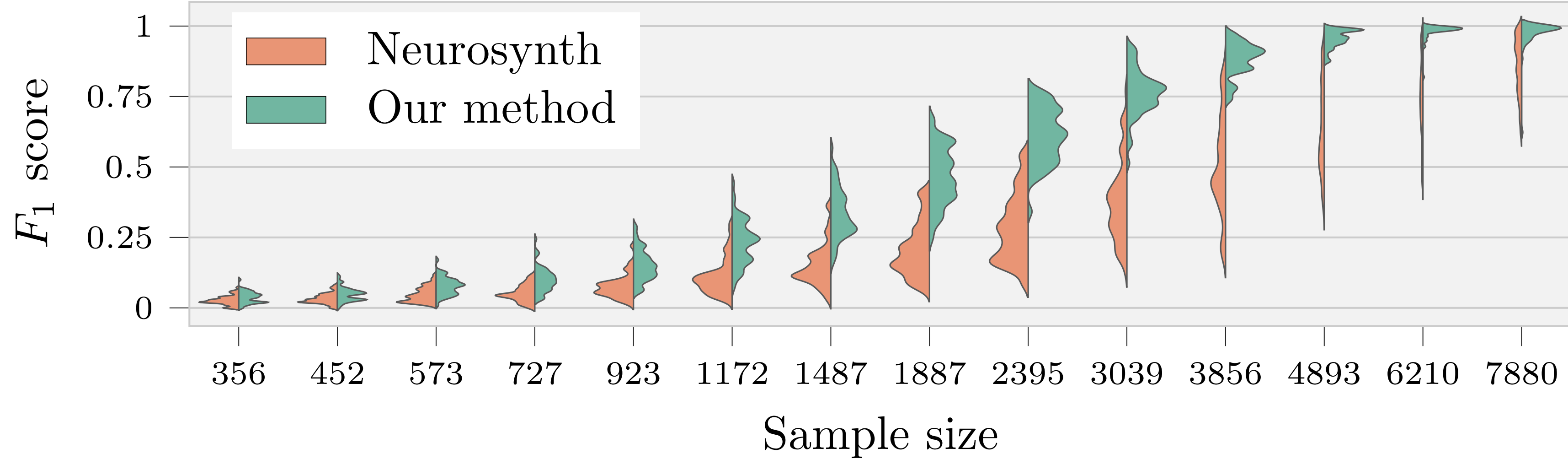
\end{document}